\documentclass{article}

\usepackage[preprint]{neurips_2026}

\usepackage[utf8]{inputenc} 
\usepackage[T1]{fontenc}    
\usepackage{hyperref}       
\usepackage{url}            
\usepackage{booktabs}       
\usepackage{amsfonts}       
\usepackage{nicefrac}       
\usepackage{microtype}      
\usepackage{xcolor}         

\usepackage{algorithm}
\usepackage{algorithmic}
\usepackage{microtype}
\usepackage{graphicx}
\usepackage{subcaption}
\usepackage{booktabs}
\usepackage{amsmath}
\usepackage{amssymb}
\usepackage{mathtools}
\usepackage{amsthm}
\usepackage{multirow}
\usepackage{wrapfig}
\usepackage{multicol}
\usepackage{enumitem}
\usepackage{caption}
\usepackage{bm}
\makeatletter
\def\blfootnote{\gdef\@thefnmark{}\@footnotetext}
\makeatother
\title{Quantized Evolution Strategies: High-precision \\Fine-tuning of Quantized LLMs at Low-precision Cost}

\author{%
  Yinggan Xu \\
  University of California, Los Angeles\\
  \texttt{yingganxu@gmail.com} \\
  \And
  Kajetan Schweighofer \\
  Cognizant AI Lab \\
  \texttt{kai.schweighofer@gmx.at} \\
  \AND
  Risto Miikkulainen \\
  UT Austin \& Cognizant AI Lab \\
  \texttt{risto@cs.utexas.edu} \\
  \And
  Xin Qiu \\
  Cognizant AI Lab \\
  \texttt{qiuxin.nju@gmail.com} \\
}

\begin{document}

\maketitle
\blfootnote{Correspondence to: Xin Qiu \textless qiuxin.nju@gmail.com\textgreater, Yinggan Xu \textless yingganxu@gmail.com\textgreater}

\captionsetup[algorithm]{font=small}

\begin{abstract}
Post-Training Quantization (PTQ) is essential for deploying Large Language Models (LLMs) on memory-constrained devices, yet it renders models static and difficult to fine-tune. Standard fine-tuning paradigms, including Reinforcement Learning (RL), fundamentally rely on backpropagation and continuous weights to compute gradients. Thus they cannot be used on quantized models, where the parameter space is discrete and non-differentiable.
While Evolution Strategies (ES) offer a backpropagation-free alternative, optimization of the quantized parameters can still fail due to vanishing or inaccurate gradient estimation.
This paper introduces Quantized Evolution Strategies (QES), an optimization paradigm that performs full-parameter fine-tuning directly in the quantized space.
QES is based on two innovations: (1) it integrates accumulated error feedback to preserve high-precision weight updating signals, and (2) it utilizes a stateless seed replay to reduce memory usage to low-precision inference levels. QES significantly outperforms the state-of-the-art zeroth-order fine-tuning methods 
on a variety of tasks, making direct fine-tuning for quantized models possible. It therefore opens up the possibility for scaling up LLMs entirely in the quantized space. The source code is available at \href{https://github.com/dibbla/Quantized-Evolution-Strategies}{https://github.com/dibbla/Quantized-Evolution-Strategies}.
\end{abstract}

\section{Introduction}
The scaling of Large Language Models (LLMs) has unlocked emergent capabilities in mathematical reasoning, coding, and general problem-solving \citep{deepseekmath, openai2024openaio1card, Guo_2025}. However, this performance comes at a significant computational cost. To mitigate the memory bottleneck of large-scale deployment, Post-Training Quantization (PTQ) has become standard practice. Techniques such as activation smoothing \citep[e.g.\ SmoothQuant,][]{xiao2024smoothquantaccurateefficientposttraining} and layer-wise weight compression (e.g.\ GPTQ; \citealp{frantar2023gptqaccurateposttrainingquantization}; AWQ; \citealp{lin2024awqactivationawareweightquantization}) make 3-4 bit precision inference possible with negligible performance degradation. It is then possible to deploy such models on consumer-grade hardware.

Whereas PTQ democratizes \textit{inference}, the model becomes essentially a static artifact. Any fine-tuning is generally done before the model is quantized, and requires massive, training-grade compute clusters that PTQ practitioners typically do not have access to. Methods that allow fine-tuning quantized models directly are limited. 
While Zeroth-Order (ZO) optimization and Evolution Strategies (ES) have emerged as memory-efficient alternatives to pre-quantization fine-tuning \citep{malladi2024finetuninglanguagemodelsjust, qiu2025evolutionstrategiesscalellm}, they are severely limited in discrete parameter spaces. Backpropagation-free approaches like QuZO and others \citep{zhou2025quzo,feng2024stepping} have shown promise in Supervised Fine-Tuning (SFT) on quantized models, but struggle with reasoning tasks. The problem, identified in this paper as the \textit{stagnation problem}, is that the discrete nature of the quantized parameter space causes gradient signals to vanish, leading to a collapse of the optimization process. Moreover, inaccurate gradient signals due to the discretization errors severely reduce the optimization efficiency.

\begin{figure}
    \hspace{4pt}
    \begin{minipage}{0.43\textwidth}
    \includegraphics[width=1\linewidth]{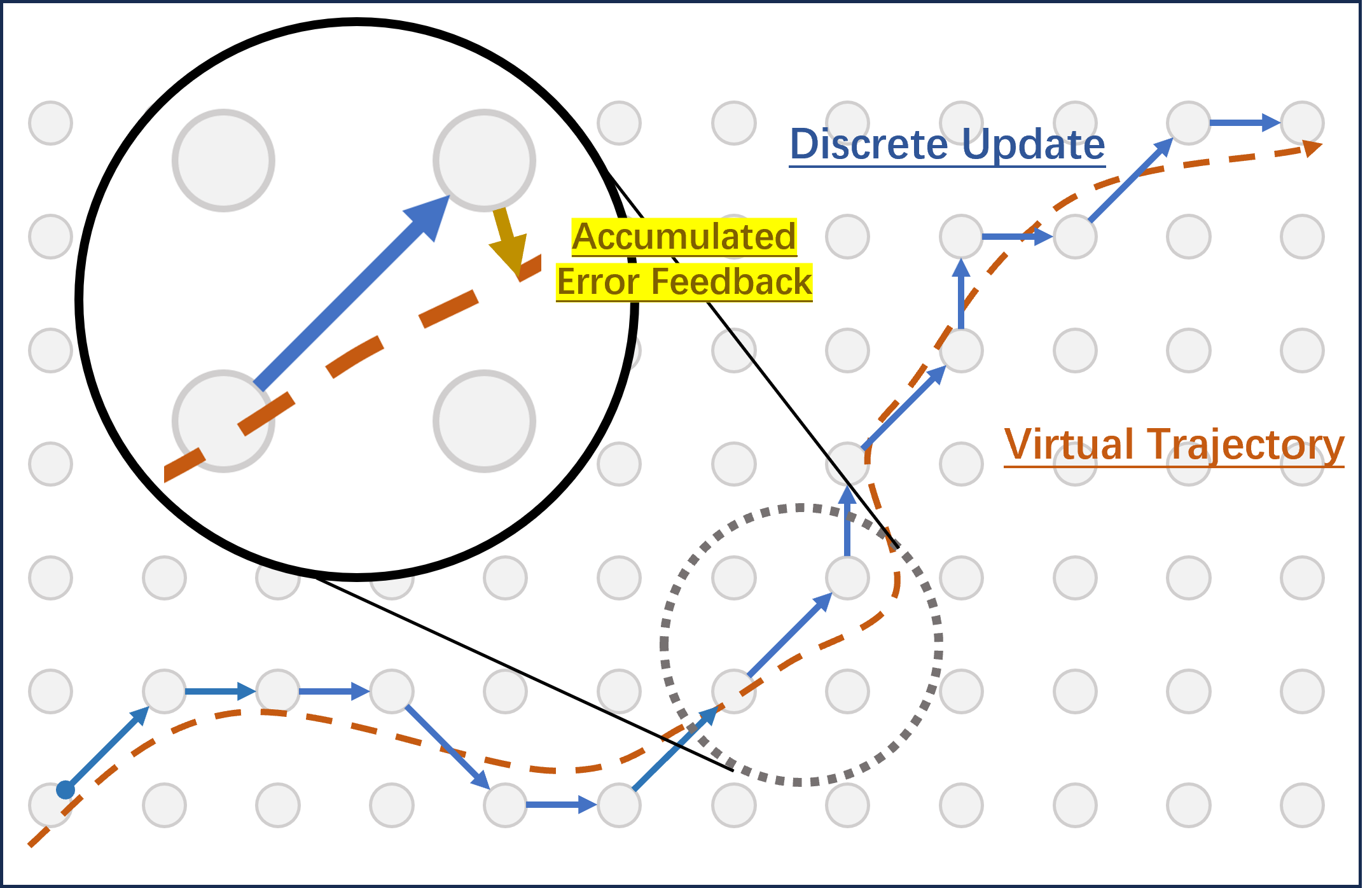}
    \end{minipage}
    \hspace{12pt}
    \begin{minipage}{0.5\textwidth}
    \captionof{figure}{An overview of Quantized Evolution Strategies (QES). The goal is to optimize quantized LLMs directly on discrete parameter space in a memory-efficient manner.  Error residuals are accumulated across iterations until they reach a threshold for making a discrete change. QES achieves temporal equivalence to the high-precision optimization trajectory, while maintaining the memory requirements of inference-only quantized models.}
    \label{fig:overview}
    \end{minipage}
\end{figure}

This paper introduces Quantized Evolution Strategies (QES; Figure~\ref{fig:overview}), a novel optimization paradigm designed to perform full-parameter fine-tuning directly on the quantized model. By bridging quantized evolution with signal processing principles, specifically Delta-Sigma modulation \citep{delta-sigma, razavi2016delta}, it proposes an accumulated error feedback mechanism that allows the optimizer to sense and traverse high-precision update trajectories even in ultra-low bit settings (e.g., four-bit integers denoted as INT4). This mechanism mirrors historical techniques in communication-efficient training, such as 1-bit stochastic gradient descent \citep{seide20141, Strom2015error-feedback, karimireddy2019error}, but adapts them for the first time to the strict constraints of backpropagation-free quantized optimization. Furthermore, to address the memory overhead of tracking these error residuals, the \textit{Stateless Seed Replay} mechanism is introduced. It reconstructs optimization states on-the-fly, reducing GPU memory requirements of QES to inference level of the low-precision quantized model.

The paper makes three main contributions:
\begin{enumerate}[leftmargin=15pt]
    \item \textbf{A Mechanism for Optimization in Quantized Space:} QES enables fine-tuning of LLMs directly within their low-precision quantized weights, yet maintains high-precision learning dynamics. Unlike gradient-based methods that require differentiable operations, QES utilizes accumulated error feedback to ensure progress on low-precision non-differentiable landscapes.
    \item \textbf{Inference-Level Memory Footprint:} By replacing the storage of high-precision optimizer states with stateless seed replay, QES reduces GPU memory consumption to the order of low-precision inference. This level allows full-parameter learning on hardware that previously could only accommodate quantized inference.
    \item \textbf{Overcoming Gradient Stagnation and Discretization inaccuracy:} QES resolves two challenges for optimization in quantized space: (1) stagnation of learning due to small gradient signals, and (2) inaccurate parameter updates due to discretization errors.
\end{enumerate}

Evaluated in two reasoning tasks and four SFT benchmarks, QES significantly improves the quantized baseline models, and does so significantly better than the state-of-the-art quantized fine-tuning method QuZO. It therefore democratizes fine-tuning, making it possible in low resource environments. The results also suggest a new avenue for scaling LLMs in the future by including more quantized parameters in the same amount of memory.

\section{Related Work}

In prior work, several methods have been developed for quantizing LLMs, for fine-tuning them without gradients, and for reducing errors resulting from quantization. These methods are reviewed in this section.

\subsection{LLM Quantization}

To mitigate the memory bottleneck of large-scale deployment, Post-Training Quantization (PTQ) has become standard practice. Techniques such as activation smoothing \citep[e.g., SmoothQuant;][]{xiao2024smoothquantaccurateefficientposttraining}) and layer-wise weight compression (GPTQ; \citealp{frantar2023gptqaccurateposttrainingquantization}; AWQ; \citealp{lin2024awqactivationawareweightquantization}) enable 3-4 bit precision inference with negligible  degradation in performance.
Quantization-aware training (QAT) approaches aim at training the high-precision network such that it is easily quantizable afterwards \citep{liu2024llm}. 

There are also works on fine-tuning the quantized models through backpropagation, e.g. by use of the straight-through estimator (STE) \citep{bengio2013estimating}. This suffers from two critical limitations: (1) STE is inherently biased, and (2) it requires full-precision latent weights and optimizer states, undermining the memory benefit of quantized models. Another approach is QLoRA \citep{dettmers2023qlora}, which freezes the quantized weights and optimizes small high-precision adapters; while memory-efficient, this restricts updates to a low-rank subspace and leaves the quantized parameters themselves unchanged. Therefore, optimization paradigms capable of performing full-parameter fine-tuning directly on the quantized model are needed.

\subsection{Zeroth Order Fine-tuning for Quantized Models}
Improving alignment and reasoning ability remains the central focus of LLM fine-tuning.
The standard approach is Reinforcement Learning with Human Feedback (RLHF), and it has been successful in general preference alignment \citep{ouyang2022traininglanguagemodelsfollow}, mathematical reasoning \citep{deepseekmath, dapo, drgrpo}, code generation \citep{gehring2024rlef, Guo_2025}, and general reasoning \citep{openai2024openaio1card, yang2025qwen3technicalreport}. However, RLHF relies fundamentally on backpropagation, restricting its application to differentiable, full-precision architectures with significant memory.

Recently, Zeroth-Order (ZO) optimizer and Evolution Strategies (ES) have emerged as backpropagation-free alternatives. Building on simultaneous perturbation stochastic approximation (SPSA; \citealp{SPSA}), the MeZO method \citep{malladi2024finetuninglanguagemodelsjust} pioneered the scaling of ZO-SGD \citep{ZO-SGD} to billion-parameter models. Similarly, recent work has demonstrated that ES can match or exceed RL performance in similar high-dimensional spaces \citep{qiu2025evolutionstrategiesscalellm}. However, these methods were originally designed for high-precision continuous parameters, and cannot be directly used on quantized models. 

Nevertheless, the backpropagation-free paradigm offers a natural basis for quantized fine-tuning. Recent approaches like QuZO \citep{zhou2025quzo} and QZO \citep{feng2024stepping} extended ZO optimizers to supervised fine-tuning of quantized models. However, so far their success in reasoning tasks has been limited; they struggle to capture sparse signals, often leading to no meaningful convergence. QES addresses this gap by connecting quantized evolution to signal processing principles.

\subsection{Techniques for Reducing Discretization Errors}

The fundamental challenge of approximating a continuous signal with discrete steps is well-studied in the context of Delta-Sigma modulation \citep{delta-sigma, razavi2016delta}. In this framework, the error introduced by taking a discrete step (i.e.\ quantization) is not discarded but accumulated and applied to subsequent steps. This mechanism, known as noise shaping in DAC design, ensures that while individual steps may be coarse, the time-averaged behavior of the system tracks the continuous trajectory accurately.

In the domain of gradient compression, this classical principle can be instantiated as residual accumulation, or error feedback \citep{seide20141, Strom2015error-feedback, karimireddy2019error}. These methods compress gradient information for efficiency but accumulate the resulting error for correction in later steps. Unlike Stochastic Rounding \citep{gupta2015deep, zhou2025quzo}, which achieves unbiased updates via probabilistic sampling but suffers from high variance, this method ensures convergence via deterministic accumulation of residuals. This process effectively "shapes" the quantization noise of the optimization trajectory, making it possible to apply quantized evolution to fine-tuning without the instability of prior methods.

\section{Method}
The challenge of quantized optimization is described first, followed by the QES method with accumulated error feedback and seed replay.

\subsection{The Optimization Challenge}
In LLM reasoning tasks, the objective is to select model parameters $\mathbf{W}$ that maximize a reward function $J(\mathbf{W})$ defined for this task: 
\begin{equation}
    \max_{\mathbf{W} \in \mathcal{W}} \mathbb{E}[J(\mathbf{W})].
\end{equation}
Standard first-order optimization methods (e.g., SGD, Adam) cannot be applied in the quantized setting for two reasons: (1) the quantization operator is non-differentiable, and (2) there is insufficient memory for storing high-precision gradients and optimizer states in many edge applications. Consequently, ES is used to estimate descent directions via parameter-space exploration.

In the standard continuous setting \citep{salimans2017evolutionstrategiesscalablealternative, qiu2025evolutionstrategiesscalellm}, ES approximates the gradient of the expected reward $\nabla \mathbb{E}[J(\mathbf{W})]$ using a population of $N$ perturbations by optimizing a Gaussian-smoothed objective $\max_{\mathbf{W} \in \mathcal{W}} \mathbb{E}_{\epsilon\in{\mathcal{N}(0,\bm{I})}}[J(\mathbf{W}+\sigma\epsilon)]$.

For a current parameter state $\mathbf{W}_t$, the gradient estimate $\hat{\boldsymbol{g}}$ is computed as:
\begin{equation} \label{eq:es_estimator}
    \hat{\boldsymbol{g}} = \frac{1}{N \sigma} \sum_{i=1}^N F_i \cdot \boldsymbol{\epsilon}_i,
\end{equation}
where $\boldsymbol{\epsilon}_i \sim \mathcal{N}(0, \bm{I})$ is a random perturbation sampled from a standard normal distribution, $\sigma$ is the standard deviation for scaling the perturbation, and $F_i=F(\mathbf{W}_t + \sigma \boldsymbol{\epsilon}_i)$ is the fitness (reward) corresponding to the perturbed weights. Note that fitness is thus the normalized reward score derived from $J(\mathbf{W}_t + \sigma \boldsymbol{\epsilon}_i)$ to make sure optimization is stable. Fitness is evaluated on a problem set according to the reinforcement learning from verifiable rewards (RLVR) framework \citep{deepseekmath}, where the model is queried and assessed based on whether its responses are correct.

Standard ES updates the parameters via gradient ascent: $\mathbf{W}_{t+1} \leftarrow \mathbf{W}_t + \alpha \hat{\boldsymbol{g}}$ with learning rate $\alpha$. While effective in continuous spaces at large scale \citep{qiu2025evolutionstrategiesscalellm}, this update rule fails in quantized spaces due to the vanishing magnitude of the update steps relative to the discretization granularity, or errors introduced during discretization. The solution is to provide error feedback in an accumulated manner, as will be described next.

\subsection{Accumulated Error Feedback}
\label{sc:accumulation}
QES optimizes quantized LLMs directly within the discrete integer space $\mathbf{W} \in \{0, \dots, 2^B-1\}^d$, where $d$ is the number of model parameters and $B$ is the number of bits to represent each parameter, Thus, the parameters $\mathbf{W}$ lie on a low-precision lattice $\mathcal{W}_{\mathcal{Q}}$ (e.g., INT4) defined by the quantization operator $\mathcal{Q}(\cdot)$. Since the continuous perturbation $\sigma \boldsymbol{\epsilon}$ from Eq.~\eqref{eq:es_estimator} violates quantization constraints, a stochastic perturbation strategy is adapted from prior work \citep{bernoulli_quant, zhou2025quzo}. 

A discrete perturbation $\boldsymbol{\delta}$ is created  by stochastically rounding the scaled Gaussian noise $\sigma \boldsymbol{\epsilon}$ (where $\boldsymbol{\epsilon} \sim \mathcal{N}(0, \bm{I})$):
\begin{equation}
\label{eq:bernouli_perturb}
    \boldsymbol{\delta} = \lfloor \sigma \boldsymbol{\epsilon} \rfloor + \boldsymbol{b}, \quad \text{where } \boldsymbol{b} \sim \text{Bernoulli}(\sigma \boldsymbol{\epsilon} - \lfloor \sigma \boldsymbol{\epsilon} \rfloor).
\end{equation}

By recording the random seeds used to generate $\boldsymbol{\epsilon}$ and $\boldsymbol{b}$, the $\boldsymbol{\delta}$ can be reproduced during optimization without the memory overhead of stored perturbation vectors \citep{qiu2025evolutionstrategiesscalellm, salimans2017evolutionstrategiesscalablealternative}.
To strictly enforce the codebook limits defined by $\mathcal{W}_{\mathcal{Q}}$, boundary gating is applied to mask invalid updates:
\begin{equation}
    \mathbf{W}'_{ij} =
        \begin{cases}
            \mathbf{W}_{ij} + \boldsymbol{\delta}_{ij} & \text{if } 0 \le \mathbf{W}_{ij} + \boldsymbol{\delta}_{ij} < 2^B, \\ \mathbf{W}_{ij} & \text{otherwise}.
        \end{cases}
\end{equation}

The gradient direction $\hat{g}$ is then estimated by aggregating the discrete search directions weighted by their rewards:
\begin{equation} 
\label{eq:grad_est}
    \hat{\boldsymbol{g}} = \frac{1}{N \sigma} \sum_{i=1}^N F_i \cdot \boldsymbol{\delta}_i.
\end{equation}
A critical challenge can be seen in Equation~\ref{eq:grad_est}: The scaled update step $\alpha \hat{\boldsymbol{g}}$ is often smaller than the minimum discretization gap of the parameter lattice (i.e., $\|\alpha \hat{\boldsymbol{g}}\|_\infty < 1$ for integer weights). Naively rounding this update either results in $\Delta \mathbf{W} = \mathbf{0}$, causing optimization to stagnate, or inaccurate weight updating signal, reducing optimization efficiency.

Interestingly, this challenge is well known in signal processing. The standard solution, Delta-Sigma modulation \citep{delta-sigma, razavi2016delta}, employs a feedback loop to accumulate quantization error over time, preserving signal fidelity. This approach is already used in communication-efficient training, such as 1-bit SGD, where uncompressed errors are carried forward to ensure convergence \citep{seide20141, Strom2015error-feedback, karimireddy2019error}.  It can be used to solve the stagnation problem and preserve weight updating accuracy in QES. 

\begin{figure}[t]
\begin{minipage}[!t]{0.49\textwidth}
\begin{algorithm}[H]
   \small
   \caption{QES with Accumulated Error Feedback}
   \label{alg:qes_ef}
   \begin{algorithmic}[1]
   \STATE {\bfseries Input:} Integer Weights $\mathbf{W}_0 \in \{0, \dots, 2^B-1\}^d$, Learning Rate $\alpha$, Decay $\gamma$, Population $N$, Maximum Iteration Number $T$
   \STATE {\bfseries Output:} Final model weights $\mathbf{W}_T$
   \STATE {\bfseries Initialize:} Residuals $\boldsymbol{e}_0 \leftarrow \mathbf{0}$ (FP16)
   \vspace{0.08cm}
   
   \FOR{$t=0$ {\bfseries to} $T-1$}
      \FOR{each perturbation $i \in \{1, \dots, N\}$ in parallel}
          \STATE Reconstruct $\boldsymbol{\delta}_i$ using Eq.~\eqref{eq:bernouli_perturb} and seed $s_i$
          \STATE Apply perturbation: $\mathbf{W}' \leftarrow \text{Gate}(\mathbf{W}_t + \boldsymbol{\delta}_i)$
          \STATE Execute inference and compute reward $F_i$
      \ENDFOR
      
      \STATE Normalize reward for population
      \STATE Estimate gradient $\hat{\boldsymbol{g}}_t \leftarrow \frac{1}{N\sigma} \sum_{i=1}^N F_i \cdot \boldsymbol{\delta}_i$
      
      \STATE $\boldsymbol{u}_t \leftarrow \alpha \hat{\boldsymbol{g}}_t + \gamma \boldsymbol{e}_t$ \quad \textit{// Apply accum. error}
      \STATE $\Delta \mathbf{W}_t \leftarrow \text{Round}(\boldsymbol{u}_t)$ \quad \textit{// Discretize update}
      \STATE $\boldsymbol{e}_{t+1} \leftarrow \boldsymbol{u}_t - \Delta \mathbf{W}_t$ \quad \textit{// Update accum. error}
      \STATE $\mathbf{W}_{t+1} \leftarrow \mathbf{W}_t + \Delta \mathbf{W}_t$ \quad \textit{// Update weight}
   \ENDFOR 
   \STATE \textbf{return} $\mathbf{W}_{T}$
   \end{algorithmic}
\end{algorithm}
\end{minipage}
\hfill
\begin{minipage}[!t]{0.48\textwidth}
\begin{algorithm}[H]
   \small
   \caption{Stateless QES Update with Seed Replay}
   \label{alg:seed_replay}
   \begin{algorithmic}[1]
   \STATE {\bfseries Input:} Current Weights $\mathbf{W}_{t}$, New Seeds $\mathcal{S}_t$, New Rewards $\mathcal{F}_t$, History $\mathcal{H}$, Window $K$, Sigma $\sigma$
   \STATE {\bfseries Output:} Updated $\mathbf{W}_{t+1}$, New History $\mathcal{H}'$

   \STATE {\bfseries Initialize:} proxy residual $\tilde{\boldsymbol{e}} \leftarrow \mathbf{0}$
   \FOR{$(\hat{\mathcal{S}}, \hat{\mathcal{F}}) \in \mathcal{H}$}
      \STATE Re-generate noise with history seed $\epsilon \leftarrow \text{RNG}(\hat{\mathcal{S}})$
      \STATE Re-compute grad $\hat{\boldsymbol{g}} \leftarrow \text{Agg}(\boldsymbol{\epsilon}, \hat{\mathcal{F}}, \boldsymbol{\sigma})$
      \STATE $\boldsymbol{u}_{sim} \leftarrow \alpha \hat{\boldsymbol{g}} + \gamma \tilde{\boldsymbol{e}}$
      \STATE $\Delta_{sim} \leftarrow \text{Round}(\boldsymbol{u}_{sim})$
      \STATE $\Delta_{final} \leftarrow \text{Gate}(\mathbf{W}_t + \Delta_{sim})$ 
      \STATE $\tilde{\boldsymbol{e}} \leftarrow \boldsymbol{u}_{sim} - \Delta_{final}$ \quad \textit{// Update proxy error}
   \ENDFOR

   \STATE Generate current noise $\boldsymbol{\epsilon}_t \leftarrow \text{RNG}(\mathcal{S}_t)$
   \STATE $\hat{\boldsymbol{g}}_t \leftarrow \text{Agg}(\boldsymbol{\epsilon}_t, \mathcal{F}_t, \sigma)$
   \STATE $\boldsymbol{u}_t \leftarrow \alpha \hat{\boldsymbol{g}}_t + \gamma \tilde{\boldsymbol{e}}$ \quad \textit{// Add rematerialized error}

   \STATE $\Delta \mathbf{W} \leftarrow \text{Round}(\boldsymbol{u}_t)$
   \STATE $\mathbf{W}_{t+1} \leftarrow \text{Gate}(\mathbf{W}_{t} + \Delta \mathbf{W})$

   \STATE $\mathcal{H}' \leftarrow \text{Enqueue}(\mathcal{H}, (\mathcal{S}_t, \mathcal{F}_t))$
   \STATE \textbf{return} $\mathbf{W}_{t+1}, \mathcal{H}'$
   \end{algorithmic}
\end{algorithm}
\end{minipage}
\end{figure}

In QES, a high-precision error vector $\boldsymbol{e}_t$ (typically FP16) that accumulates the quantization error from previous steps is maintained. Rather than discarding the fractional component of the update, it is carried forward. The update dynamics at step $t$ are thus defined as:
\begin{align}
    \boldsymbol{u}_t &= \alpha \hat{\boldsymbol{g}}_t + \gamma \boldsymbol{e}_{t-1} \label{eq:accum} \\
    \Delta \mathbf{W}_t &= \text{Round}(\boldsymbol{u}_t) \\
    \boldsymbol{e}_t &= \boldsymbol{u}_t - \Delta \mathbf{W}_t \label{eq:resid},
\end{align}
where $\boldsymbol{u}_t$ represents the desired high-precision update, and $\gamma \in (0, 1]$ is a decay factor that stabilizes the history. This mechanism allows infinitesimal gradient signals to accumulate in $\boldsymbol{e}$ over multiple iterations until they cross the rounding threshold ($|\boldsymbol{u}| \ge 0.5$), effectively simulating a lower learning rate on the integer lattice. The full quantized evolution strategy with accumulated error feedback is listed in Algorithm \ref{alg:qes_ef}.

\subsection{Stateless Seed Replay}

While Algorithm \ref{alg:qes_ef} ensures progress, it incurs a significant memory penalty. Storing the dense high-precision error vector $\boldsymbol{e}_t \in \mathbb{R}^d$ (typically FP16) often consumes more VRAM than the quantized model weights themselves, canceling the primary advantage of quantization.

To resolve this bottleneck, QES includes \textit{Stateless Seed Replay}. Note that the error state $\boldsymbol{e}_t$ is deterministic given the initial condition and the history of optimization steps. Rather than persisting $\boldsymbol{e}_t$, it can be \textit{rematerialized} on-the-fly by replaying a limited history window. 
To do that, a lightweight history buffer $\mathcal{H}_t = \{(\mathcal{S}_\tau, \mathcal{F}_\tau)\}_{\tau=t-K}^{t-1}$ is maintained, containing only the random seeds $\mathcal{S}$ and the scalar rewards $\mathcal{F}$ for the past $K$ steps' populations. To perform an update at step $t$, the error accumulation process is re-simulated starting from an assumed zero error state at step $t-K$. Since the decay factor $\gamma \in (0, 1)$, the contribution of errors from steps $\tau < t-K$ vanishes exponentially ($\gamma^K \approx 0$). 

During the replay of step $\tau$, boundary constraints are checked using the current weights $\mathbf{W}_t$ rather than reconstructing the historical weights $\mathbf{W}_\tau$. Since discrete updates are sparse, $\mathbf{W}_\tau \approx \mathbf{W}_t$, and the discrepancy in boundary masking is minimal.

This approach, detailed in Algorithm \ref{alg:seed_replay}, trades computation for memory. By performing $K$ additional reconstruction operations per update, the optimizer state memory complexity is reduced from $O(d)$ to $O(K)$. This tradeoff is highly favorable because $K$ is orders of magnitude smaller than the parameter dimension $d$. In a typical LLM optimization scenario, $d$ may exceed $10^9$, whereas a short window of $K \approx 50$ is sufficient to recover the accumulated error. For standard values (e.g., $\gamma=0.9, K=50$), the influence introduced by truncating history is negligible.

\section{Experiments}
\label{sec:experiments}

QES is evaluated on three task families of increasing difficulty: classification-style supervised fine-tuning (SFT), arithmetic reasoning, and grade-school mathematics. The goal is to demonstrate that QES fine-tunes LLMs directly in their quantized form---without full-precision gradients or auxiliary optimizer states---across both differentiable and non-differentiable objectives. Section~\ref{subsec:setup} describes the setup; Sections~\ref{subsec:sft}--\ref{subsec:reasoning} report main results; Sections~\ref{subsec:ablation}--\ref{subsec:accel} analyze the residual and seed-replay mechanisms.

\subsection{Experimental Setup}
\label{subsec:setup}

\paragraph{Tasks.}
RoBERTa-large is fine-tuned on the \textbf{SFT} tasks SNLI, MNLI, RTE, and SST-5 following the protocol of MeZO~\citep{malladi2024finetuninglanguagemodelsjust}; performance is reported as classification accuracy. \textbf{Countdown}~\citep{gandhi2024stream, tinyzero}: a compact reasoning task in which the model must produce an arithmetic expression over $\{+,-,\times,/\}$ that evaluates to a target value (e.g., from $\{3,4,28,52\}$ targeting $44$, the solution is $28+52/4+3$). \textbf{GSM8K}~\citep{cobbe2021training}: grade-school math word problems requiring multi-step numerical reasoning. For both reasoning tasks the prompt template from GRPO-Zero~\citep{policygradient2025grpozero} is used and the reward is binary correctness.

\paragraph{Models and quantization.}
Qwen2.5-1.5B and -3B~\citep{qwen2025qwen25technicalreport} serve as the reasoning backbones, quantized to INT4 and INT8 with GPTQ~\citep{frantar2023gptqaccurateposttrainingquantization} and to W8A8 (both weights and activations are quantized to INT8 precision) with LLM-Compressor~\citep{llmcompressor2024}. RoBERTa-large is used at FP32 and at W8 (GPTQ). We also tested the larger scale models in Appendix~\ref{app:llama8b}.


\paragraph{Baselines.}
QES is compared against: the \textbf{Base Model} (quantized, no fine-tuning); \textbf{QuZO}~\citep{zhou2025quzo}, a quantized zeroth-order method that is the primary quantized baseline in both regimes; \textbf{MeZO}~\citep{malladi2024finetuninglanguagemodelsjust}, full-precision zeroth-order SGD (not applicable to quantized space); \textbf{First-Order (FO)} backpropagation, in FP32 (upper bound) and on the W8 backbone with straight-through estimation (SFT only); and \textbf{Full Residual}, a QES variant that stores FP16 residuals, used as an oracle in the residual ablation (Section~\ref{subsec:ablation}).

We specify more details of training and implementations in Appendix~\ref{app:hyperparameters}.

\subsection{Supervised Fine-Tuning}
\label{subsec:sft}

Table~\ref{tab:sft_results} reports SFT accuracy on the four classification tasks. Among methods constrained to the W8 backbone, QES improves substantially over QuZO ($44.4$ vs.\ $34.2$ on average) and over the W8 first-order baseline with STE($41.0$). Notably, QES also outperforms full-precision MeZO ($36.5$), suggesting that error feedback recovers more useful update signal than smoothed ZO estimation, even when the latter operates in FP32. The remaining gap to full-precision first-order ($57.3$) reflects the cost of optimizing on a discrete lattice with only forward-pass evaluations; QES closes most of that gap while requiring no backpropagation and no optimizer state.

\begin{table}[!t]
\centering
\setlength{\tabcolsep}{11.2pt}
\caption{SFT accuracy (\%) on RoBERTa-large. Best W8 result in each column is in bold. QES achieves the best performance among quantized methods and on several tasks surpasses the W8 first-order baseline and the FP32 MeZO zeroth-order baseline.}
\label{tab:sft_results}
\vskip 0.05in
\begin{small}
\begin{sc}
\begin{tabular}{lllcccc|c}
\toprule
\textbf{Method} && \textbf{Prec.} & \textbf{SNLI} & \textbf{MNLI} & \textbf{RTE} & \textbf{SST-5} & \textbf{Avg} \\
\midrule
\multicolumn{8}{l}{\emph{Full precision (upper bound)}} \\
First-Order            && FP32 & 72.9 & 61.1 & 49.0 & 46.2 & 57.3 \\
MeZO         && FP32 & 34.0 & 34.0 & 56.2 & 21.7 & 36.5 \\
\midrule
\multicolumn{8}{l}{\emph{8-bit weights}} \\
First-Order + STE        && W8   & 50.0 & \textbf{44.4} & 49.0 & 20.4 & 41.0 \\
QuZO           && W8   & 32.3 & 40.3 & 44.8 & 19.6 & 34.2 \\
\textbf{QES (ours)} && W8 & \textbf{55.6} & 42.4 & \textbf{55.2} & \textbf{24.4} & \textbf{44.4} \\
\bottomrule
\end{tabular}
\end{sc}
\end{small}
\end{table}

\subsection{Reasoning}
\label{subsec:reasoning}

Table~\ref{tab:main_results} summarizes accuracy on Countdown and GSM8K across model sizes and quantization formats. Two patterns are immediate. First, QuZO is brittle on the coarse INT4 lattice and on the smaller 1.5B model: on Qwen2.5-1.5B INT4, it improves over base by only $1.75$ points on Countdown and fails to attain any improvement on GSM8K. Second, QES achieves consistent and often large gains across all settings. The most striking case is Qwen2.5-3B in W8A8, where QuZO collapses to $4.40\%$ on GSM8K while QES reaches $80.82\%$. The gap between QuZO and QES widens with task difficulty (Countdown $\rightarrow$ GSM8K) and narrows with larger model scale, consistent with the intuition that smaller, more aggressively quantized models present sharper, more brittle landscapes for stateless ZO methods~\citep{li2018visualizing}. As a scaling-up case study, Appendix~\ref{app:llama8b} reports Llama-3.1-8B INT4 reaching $82.64\%$ on GSM8K from a $64.14\%$ base. Countdown training curves are shown in Figure~\ref{fig:training_dynamics} in Appendix~\ref{app:dynamics}.

\begin{table}[!t]
\centering
\setlength{\tabcolsep}{10.1pt}
\caption{Reasoning accuracy (\%) on Countdown and GSM8K across model sizes and quantization formats. QES delivers consistent and often large gains over both the quantized base and QuZO across every model size and format.}
\label{tab:main_results}
\vskip 0.05in
\begin{small}
\begin{sc}
\begin{tabular}{llcccccc}
\toprule
& & \multicolumn{3}{c}{\textbf{Countdown}} & \multicolumn{3}{c}{\textbf{GSM8K}} \\ \cmidrule(lr){3-5} \cmidrule(lr){6-8}
\textbf{Model} & \textbf{Format} & Base & QuZO & QES & Base & QuZO & QES \\
\midrule
\multirow{3}{*}{Qwen2.5-1.5B}
 & INT4 & 3.50 & 5.25  & \textbf{16.00} & 0.00  & 0.00  & \textbf{9.86}  \\
 & INT8 & 4.20 & 4.50  & \textbf{26.35} & 1.59  & 1.44  & \textbf{12.21} \\
 & W8A8 & 4.20 & 4.20  & \textbf{15.35} & 3.56  & 4.17  & \textbf{12.28} \\
\midrule
\multirow{3}{*}{Qwen2.5-3B}
 & INT4 & 2.80 & 14.25 & \textbf{31.85} & 48.45 & 48.60 & \textbf{77.56} \\
 & INT8 & 4.50 & 15.85 & \textbf{37.40} & 11.90 & 54.28 & \textbf{78.77} \\
 & W8A8 & 8.20 & 10.75 & \textbf{21.35} & 24.49 & 4.40  & \textbf{80.82} \\
\bottomrule
\end{tabular}
\end{sc}
\end{small}
\end{table}

\subsection{Ablation Studies of Stateless Seed Replay}
\label{subsec:ablation}

\paragraph{Residual storage (Stateless Replay vs.\ Full Residual).}
The Full Residual variant of QES stores FP16 residuals explicitly and serves as an oracle that isolates the contribution of error feedback when memory is unconstrained. Table~\ref{tab:residual_ablation} compares it against Stateless Seed Replay on Countdown. The two are within a few points across all six configurations and, in three of six, the seed-replay variant matches or exceeds the oracle---suggesting that any approximation introduced by replaying historical states with current weights is, in practice, on the scale of seed variance. The largest gap is Qwen2.5-3B W8A8 ($21.35$ vs.\ $31.70$); on the same configuration on GSM8K, however, the gap closes to within $0.76$ points (Full Residual: $81.58\%$, QES: $80.82\%$), indicating that this discrepancy is task- and seed-specific rather than a systematic failure of the approximation.

\paragraph{Window size and decay.}
Appendix~\ref{app:ablations} (Table~\ref{tab:window_decay} Top) evaluates the replay window $K$ and decay factor $\gamma$ on Qwen2.5-1.5B INT4 (Countdown). Two regimes are compared: $\gamma$ scaled so that historical errors vanish within the window ($\gamma^K \approx 0$), and $\gamma$ fixed at $0.90$. Under the scaled regime, performance is highly sensitive to decay: shrinking $K$ from $50$ to $10$ forces $\gamma$ down to $0.58$ and collapses accuracy from $16.00\%$ to $4.55\%$. Under fixed $\gamma=0.90$, even $K{=}10$ retains $13.05\%$, showing that the collapse is driven by aggressive decay rather than truncated history. In practice this means $K$ is a tunable knob: shrinking it from $50$ to $20$ reduces reconstruction cost by $60\%$ while losing only $1.25$ points (fixed decay).


\subsection{Fidelity of Stateless Seed Replay}
\label{subsec:fidelity}

In Stateless Seed Replay the perturbation noise $\epsilon$ is regenerated exactly from the preserved seed, but the boundary gating uses the current weights $W_t$ rather than the historical weights $W_\tau$. A reconstruction error therefore arises only when a parameter actively crosses a quantization boundary within the replay window. Define the \emph{update ratio} as the fraction of parameters changed per step, and the \emph{boundary hit ratio} $\rho$ as the fraction of those changes that land on a quantization boundary. Appendix~\ref{app:ablations} (Table~\ref{tab:window_decay}, bottom) shows the update ratio is uniformly $\approx 10^{-2}$ and $\rho$ is negligible across formats (e.g., $<10^{-5}$ for INT4). The intersection---an active update precisely at a boundary---is exceedingly rare, which is why the seed-replay variant tracks the Full-Residual oracle with near-perfect fidelity.

\subsection{Computational Cost Analysis}
\label{subsec:accel}

While Stateless Seed Replay uses negligible memory (a few KB), its compute cost scales linearly with $K$. As shown above, $K{=}20$ retains most of the accuracy of $K{=}50$ at $40\%$ of the cost, so the replay horizon trades marginal accuracy for throughput. Furthermore, since the model is not serving inference during the update phase, reconstruction can be parallelized across layers using otherwise-idle memory (e.g., offloaded KV caches). Wall-clock measurements on A100s show that, with rollouts and replay using the same number of GPUs, total training time increases by ${\sim}16.7\%$ for 1.5B and ${\sim}12.5\%$ for 3B over the no-replay oracle. A peak-VRAM comparison against Full Residual and QuZO is provided in Appendix~\ref{app:memory}.

\section{Temporal Equivalence to Continuous Optimization}
\label{sec:discussion_convergence}

To understand why QES succeeds while other quantized methods stagnate or converge slowly, it is useful to characterize QES's optimization dynamics relative to an ideal underlying continuous model. While the quantized parameters $\mathbf{W}$ are constrained to a discrete grid $\mathcal{W}_{\mathcal{Q}}$ with spacing $\Delta$, the reward function $J(\mathbf{W})$ is defined over a continuous domain. For instance, assume a Gaussian smoothed objective $J_{\sigma}(\mathbf{W}) = \mathbb{E}_{\boldsymbol{\epsilon} \sim \mathcal{N}(\mathbf{0}, \mathbf{I})}[J(\mathbf{W} + \sigma \boldsymbol{\epsilon})]$ (Figure \ref{fig:illustrate_discrete} in Appendix~\ref{app:continuous}). It can be optimized by high-precision gradient ascent, but such a mechanism is not available on the discrete grid. Instead, the gradient needs to be approximated.

One approach is to use a stochastic gradient estimator $\hat{\boldsymbol{g}}_t$, like QuZO does \cite{zhou2025quzo}. Utilizing stochastic perturbation (double quantization) yields an unbiased estimator of the smoothed gradient: $\mathbb{E}[\hat{\boldsymbol{g}}_t] = \nabla J_{\sigma}(\mathbf{W}_t)$. However, while an unbiased gradient is necessary, it is not sufficient for high performance on a discrete grid. The critical failure occurs at the application of the gradient update.

Consider the standard update rule at step $t$ with learning rate $\alpha$ and a quantization operator $\mathcal{Q}$:
\begin{equation}
    \mathbf{W}_{t+1} = \mathbf{W}_t + \mathcal{Q}(\alpha \hat{\boldsymbol{g}}_t).
\end{equation}
The quantization operator can be decomposed into the identity plus an error term $\xi$, such that $\mathcal{Q}(\boldsymbol{x}) = \boldsymbol{x} + \boldsymbol{\xi}$. Expanding the trajectory over $T$ steps thus yields
\begin{equation}
    \label{eq:quzo_decomp}
    \mathbf{W}_T = \mathbf{W}_0 + \underbrace{\sum_{t=0}^{T-1} \alpha \hat{\boldsymbol{g}}_t}_{\text{\tiny Ideal Continuous Update}} + \underbrace{\sum_{t=0}^{T-1} \boldsymbol{\xi}_t.}_{\text{\tiny Accumulated Quantization Loss}}
\end{equation}
The mechanisms behind the failure of stateless updates can be seen in this equation. First, stagnation occurs because the update magnitude in fine-tuning is often smaller than the grid precision ($\|\alpha \hat{\boldsymbol{g}}_t\|_\infty < \Delta/2$). Consequently, $\mathcal{Q}(\boldsymbol{u}) \to 0$, implying the error is $\boldsymbol{\xi}_t = - \alpha \hat{\boldsymbol{g}}_t$; thus, the \textit{\small Accumulated Quantization Loss} exactly cancels the \textit{\small Ideal Continuous Update}, resulting in $\mathbf{W}_T = \mathbf{W}_0$. Second, variance explosion arises if $\mathcal{Q}$ is stochastic. In this case, $\boldsymbol{\xi}_t$ becomes a zero-mean noise variable with standard deviation proportional to $\Delta$. These errors accumulate as a random walk, scale with $\sqrt{T}\Delta$ and create a noise floor that drowns out the subtle fine-tuning signal in the long run.

QES solves this problem by introducing a residual state $\boldsymbol{e}_t$ to enforce \textit{temporal equivalence} between the discrete and continuous domains.
%
%
First, \textit{Virtual Continuous Parameters} $\boldsymbol{\Theta}_t$ are defined as the sum of the physical discrete weights and the carry-over residual error:
\begin{equation}
\label{eq:virtualtheta}
    \boldsymbol{\Theta_t} \triangleq \mathbf{W}_t + \boldsymbol{e}_t.
\end{equation}
Then, expanding this equation with the QES update rules in Equations~\eqref{eq:accum}--\eqref{eq:resid} gives
\begin{equation}
\begin{aligned}
    \boldsymbol{\Theta}_{t+1} &= \mathbf{W}_{t+1} + \boldsymbol{e}_{t+1} \\
                 &= (\mathbf{W}_t + \Delta \mathbf{W}_t) + (\boldsymbol{u}_t - \Delta \mathbf{W}_t) \\
                 &= \mathbf{W}_t + \boldsymbol{u}_t = \mathbf{W}_t + (\alpha \hat{\boldsymbol{g}}_t + \boldsymbol{e}_t) \\
                 &= (\mathbf{W}_t + \boldsymbol{e}_t) + \alpha \hat{\boldsymbol{g}}_t = \boldsymbol{\Theta}_t + \alpha \hat{\boldsymbol{g}}_t.
\end{aligned}
\end{equation}

Thus, the virtual parameters $\boldsymbol{\Theta}_t$ evolve according to the dynamics of unconstrained, high-precision gradient ascent. The physical quantized weights $W_T$ at any step $T$ are related to this ideal trajectory by
\begin{equation}
    \mathbf{W}_T = \boldsymbol{\Theta}_T - \boldsymbol{e}_T = \left( \mathbf{W}_0 + \sum_{t=0}^{T-1} \alpha \hat{\boldsymbol{g}}_t \right) - \boldsymbol{e}_T.
\end{equation}
Crucially, unlike the stateless baseline where errors sum linearly or as a random walk, the deviation in QES is determined solely by the \textit{single final residual} $\boldsymbol{e}_T$. Since
$\boldsymbol{\Theta}_T$
rounds to the nearest grid point, this error is strictly bounded by the grid resolution: $\|\boldsymbol{e}_T\|_\infty \le \Delta/2$.

Consequently, QES guarantees that the quantized model $\mathbf{W}_t$ never deviates more than half a grid step from the ideal high-precision trajectory $\boldsymbol{\Theta}_t$. The residual $\boldsymbol{e}_t$ effectively integrates infinitesimal gradient signals over time until they cross the quantization threshold, triggering a discrete update $\Delta \mathbf{W} \neq 0$ that aligns the discrete model with the virtual continuous path. 

\section{Limitation and Future Work} \label{sec:limitation_future_work}

Several promising directions exist in overcoming the limitations of the current work and extending the capabilities of QES. First, this paper only focuses on standard linear integer quantization (INT4, INT8, W8A8). It should be possible to extend QES to more aggressive and non-uniform quantization paradigms, such as binary networks and floating-point formats (e.g., FP4). Second, the current stateless replay relies on a fixed lookback window $K$ to manage the compute-memory trade-off. An adaptive algorithm could be created that automatically tunes $K$ and the decay rate $\gamma$ based on real-time convergence stability or available hardware resources, effectively removing the need for manual hyperparameter selection.  

While the primary motivation for quantization so far has been to run existing large models on limited hardware, QES fine tuning opens up a new frontier: given the same hardware, a model with significantly more parameters can be quantized in the beginning, and QES be used to train the model directly in quantized space. Such scale-up is possible based on two aspects: (1) precision can be traded for more parameters, e.g.\ four-fold in case of INT4, and (2) only low-precision inference is needed during training, requiring about 1/12 of the memory of backpropagation \cite{malladi2024finetuninglanguagemodelsjust}. Stacking these two benefits together opens up possibilities to train one or two orders of magnitude larger models with the same hardware.

\section{Conclusion}

Several barriers make effective fine-tuning of quantized LLMs difficult: gradient stagnation, inaccurate parameter update due to discretization, and prohibitive memory requirements. This paper introduced QES, a novel backpropagation-free optimization framework designed to perform full-parameter optimization directly on quantized integer weights. First, vanishing or inaccurate gradient estimations were identified as the primary cause of failure for existing ZO methods. Inspired by Delta-Sigma modulation, an accumulated error feedback mechanism was constructed to preserve and leverage these minuscule learning signals effectively. Second, to eliminate the memory overhead of storing high-precision error states, a Stateless Seed Replay mechanism was developed. This mechanism makes full-parameter fine-tuning possible within the strict memory constraints of standard quantized inference. These mechanisms were evaluated empirically on a challenging reasoning task. QES successfully overcame the quantized gradient issues and memory restrictions, significantly outperforming the state-of-the-art quantized fine-tuning baseline and achieving parity with memory-intensive oracles. Thus, by enabling the adaptation of large models on consumer-grade hardware, QES represents a significant step toward democratizing access to LLM fine-tuning.


\bibliographystyle{unsrtnat}
\bibliography{example_paper}


\appendix

\section{Hyperparameters and Implementation Details}
\label{app:hyperparameters}

This appendix lists all hyperparameters used to produce the results in Section~\ref{sec:experiments}. Section~\ref{app:hp:common} describes settings shared across both task families; Section~\ref{app:hp:sft} covers the SFT experiments on RoBERTa-large reported in Table~\ref{tab:sft_results}; and Section~\ref{app:hp:reasoning} covers the reasoning experiments on Qwen2.5 reported in Table~\ref{tab:main_results}. Code and launch scripts to reproduce every cell are provided in the supplementary material.

\subsection{Common Settings}
\label{app:hp:common}

Across both task families, QES uses Algorithm~\ref{alg:qes_ef} with the following defaults: population size $N=8$ (for SFT) and $N=50$ (for reasoning) antithetic perturbation pairs per generation, decay factor $\gamma=0.9$, and a 4-bit stochastically rounded perturbation tensor for $\boldsymbol{\delta}$ (independent of the weight bit-width $B$). Weights are quantized with a symmetric, per-output-channel grid: scale $s = \max_{j}|\mathbf{W}_{ij}|/(2^{B-1}-1)$, range $[-(2^{B-1}-1), 2^{B-1}-1]$. Activations are kept in FP for INT4/INT8 and quantized to INT8 in W8A8 via LLM-Compressor~\citep{llmcompressor2024}. Following the LLM-QAT convention~\citep{liu2024llm}, the language-model head, classifier, and scoring projections are excluded from quantization. The Stateless Seed Replay variant maintains a circular buffer of $(\mathcal{S}_\tau, \mathcal{F}_\tau)$ tuples for the past $K$ generations; the Full Residual variant instead stores the FP16 residual $\boldsymbol{e}_t$ directly (Section~\ref{sec:experiments}).

\subsection{SFT Tasks}
\label{app:hp:sft}

We follow the prompt-based $k$-shot fine-tuning protocol of QuZO~\citep{zhou2025quzo} and MeZO~\citep{malladi2024finetuninglanguagemodelsjust}. RoBERTa-large is the backbone for all four tasks (SNLI, MNLI, RTE, SST-5); the W8 column of Table~\ref{subsec:sft} uses the same backbone post-training-quantized with GPTQ~\citep{frantar2023gptqaccurateposttrainingquantization}. We use $K_{\text{shot}}=16$ examples per class, three random seeds (42, 13, 21) for the few-shot draws, and report the mean accuracy across seeds. Optimization uses a constant learning-rate schedule (no warmup, no weight decay), batch size $8$, and maximum sequence length $128$ for SST-5 and $256$ for SNLI/MNLI/RTE. Final-checkpoint evaluation uses the LM-BFF template, verbalizer, and label-fitting pipeline.

Method-specific settings are as follows. The first-order FP32 upper bound uses AdamW at learning rate $1\!\times\!10^{-5}$. The first-order W8 baseline uses the same optimizer and learning rate, with weights snapped onto the W8 grid after each \texttt{optimizer.step()} (gradients pass through unchanged; this is equivalent to a post-step straight-through estimator). MeZO at FP32 uses $N=2$ SPSA pairs, perturbation scale $\sigma=1\!\times\!10^{-3}$, and learning rate $1\!\times\!10^{-6}$. QuZO uses $N=8$ at the per-task learning rate listed in Table~\ref{tab:sft_results}, with stochastic round-to-nearest after each step. QES uses $N=8$, $\gamma=0.9$, $\sigma=1\!\times\!10^{-3}$, deterministic per-channel snap, and the per-task learning rate $\alpha$, replay window $K$, and step budget shown in Table~\ref{tab:hp:sft}. Per-task values were tuned on seed 42 only and then frozen across the remaining two seeds.

\begin{table}[h]
\centering
\caption{Per-task QES hyperparameters for the SFT experiments on RoBERTa-large (Table~\ref{tab:sft_results} $\alpha$ is the learning rate, $K$ is the seed-replay window, and Steps is the total number of optimization steps. Population $N=8$, decay $\gamma=0.9$, and perturbation scale $\sigma=1\!\times\!10^{-3}$ for all tasks.}
\label{tab:hp:sft}
\begin{tabular}{lccc}
\toprule
\textbf{Task} & $\boldsymbol{\alpha}$ & $\boldsymbol{K}$ & \textbf{Steps} \\
\midrule
SNLI  & $3\!\times\!10^{-7}$ & 16 & 1500 \\
MNLI  & $5\!\times\!10^{-7}$ & 16 & 1500 \\
RTE   & $1\!\times\!10^{-6}$ & 16 & 1000 \\
SST-5 & $5\!\times\!10^{-7}$ & 16 & 1500 \\
\bottomrule
\end{tabular}
\end{table}

\subsection{Reasoning Tasks}
\label{app:hp:reasoning}

For Countdown and GSM8K we use Qwen2.5-1.5B and Qwen2.5-3B~\citep{qwen2025qwen25technicalreport} as backbones, post-training-quantized to INT4 and INT8 with GPTQ~\citep{frantar2023gptqaccurateposttrainingquantization} and to W8A8 with LLM-Compressor~\citep{llmcompressor2024}. The reward is binary correctness following the GRPO-Zero prompt format~\citep{policygradient2025grpozero}; rollouts are evaluated on batches of problems sampled from the training split, and accuracy is reported on a held-out 400-problem evaluation set. Each method is run for 300 generations with $N=50$ antithetic pairs per generation. The $\sigma$ and $\alpha$ values used by QES are listed in Table~\ref{tab:hp:reasoning}. The replay window is $K=50$ for the seed-replay variant on the reasoning tasks (the $K=20$ comparison in Table~\ref{tab:window_decay} is reported as an ablation), and $\gamma=0.9$ throughout. The Full Residual variant uses the same $\sigma$ and $\alpha$ as Seed Replay, with FP16 residuals replacing the seed-and-reward buffer.

\begin{table}[h]
\centering
\caption{Per-configuration QES hyperparameters for the reasoning experiments (Table~\ref{tab:main_results}). Seed Replay and Full Residual share the same $\sigma$ and $\alpha$ at each cell; the only difference between the two variants is whether residuals are stored in FP16 directly or rematerialized via seed replay. Population $N=8$, decay $\gamma=0.9$, and replay window $K=50$ throughout.}
\label{tab:hp:reasoning}
\begin{tabular}{llcc}
\toprule
\textbf{Model} & \textbf{Format} & $\boldsymbol{\sigma}$ & $\boldsymbol{\alpha}$ \\
\midrule
\multirow{3}{*}{Qwen2.5-1.5B}
 & INT4  & $1\!\times\!10^{-2}$ & $5\!\times\!10^{-4}$ \\
 & INT8  & $1\!\times\!10^{-3}$ & $1\!\times\!10^{-4}$ \\
 & W8A8  & $1\!\times\!10^{-2}$ & $1\!\times\!10^{-3}$ \\
\midrule
\multirow{3}{*}{Qwen2.5-3B}
 & INT4  & $5\!\times\!10^{-3}$ & $3\!\times\!10^{-4}$ \\
 & INT8  & $1\!\times\!10^{-3}$ & $1\!\times\!10^{-4}$ \\
 & W8A8  & $1\!\times\!10^{-2}$ & $1\!\times\!10^{-3}$ \\
\bottomrule
\end{tabular}
\end{table}
\section{Training Dynamics}
\label{app:dynamics}
\begin{figure*}[ht]
    \centering
    \includegraphics[width=0.9\linewidth]{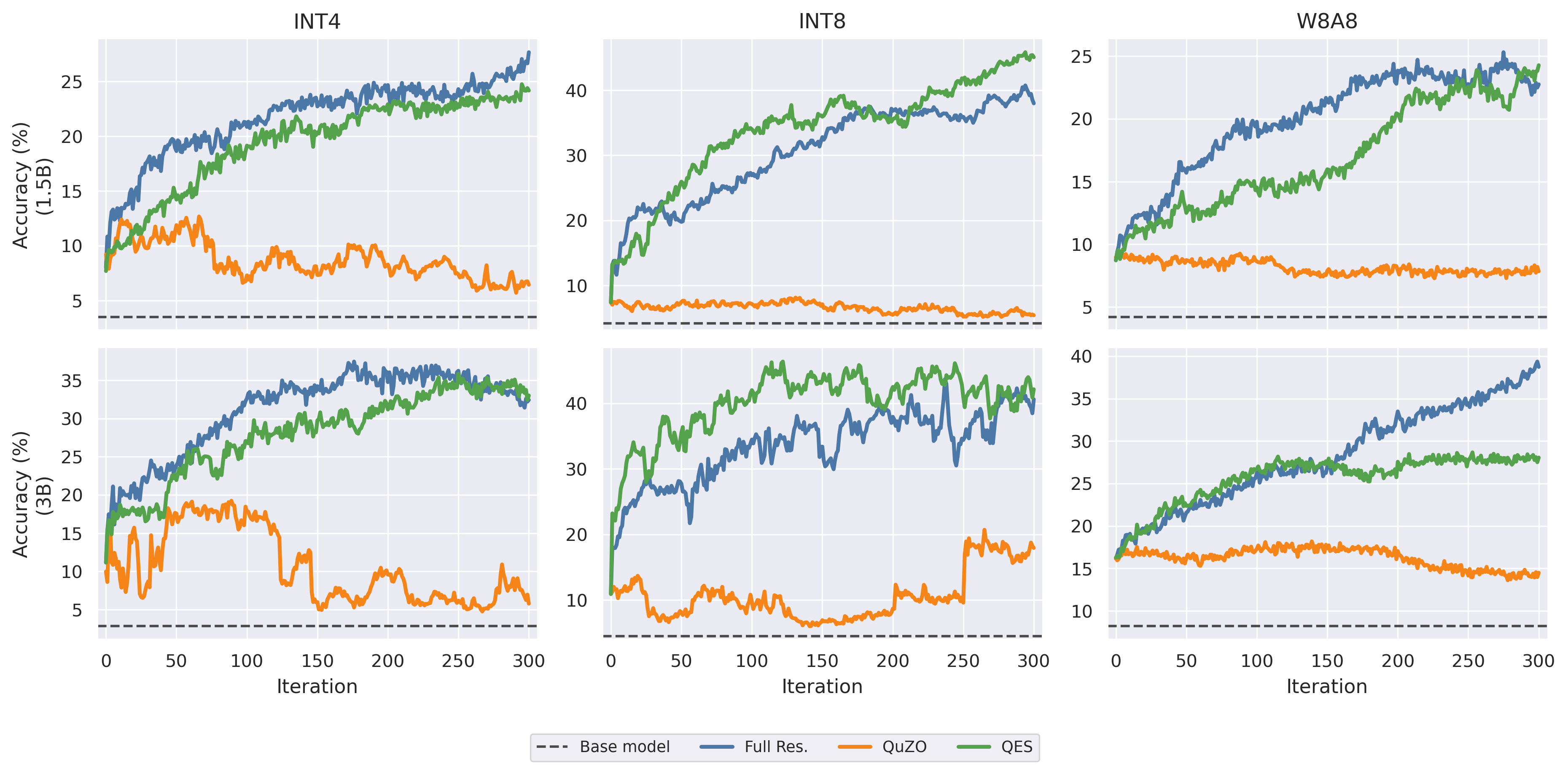}
    \caption{Countdown training curves for QuZO, QES, and Full-Residual QES against the Base Model. QuZO (orange) is unstable and frequently collapses on the INT4 lattice and on the 1.5B model. QES (green) tracks the Full-Residual oracle (blue) closely while using orders of magnitude less optimizer memory.}
    \label{fig:training_dynamics}
\end{figure*}
\section{Scaling Case Study: Llama-3.1-8B}
\label{app:llama8b}

To assess whether QES generalizes beyond the Qwen2.5 family and to model scales larger than the 1.5B/3B regime studied in the main paper, we conducted a preliminary scaling study with Llama-3.1-8B on GSM8K. The model was post-training-quantized to INT4 with GPTQ~\cite{frantar2023gptqaccurateposttrainingquantization} following the same protocol as the Qwen2.5 backbones in Section~\ref{subsec:setup}. Crucially, QES used the same fixed configuration as the Qwen2.5-3B (INT4) reasoning experiments---no per-model hyperparameter search was performed.

\begin{table}[ht]
\centering
\caption{GSM8K accuracy (\%) on Llama-3.1-8B (INT4). QES reuses the Qwen2.5-3B (INT4) reasoning hyperparameters verbatim; no per-model tuning was performed.}
\label{tab:llama8b}
\vskip 0.05in
\begin{small}
\begin{sc}
\begin{tabular}{lcc}
\toprule
\textbf{Model} & \textbf{Base} & \textbf{QES} \\
\midrule
Llama-3.1-8B (INT4) & 64.14 & \textbf{82.64} \\
\bottomrule
\end{tabular}
\end{sc}
\end{small}
\end{table}

QES improves accuracy from $64.14\%$ to $82.64\%$, an $18.5$-point gain. The result is consistent across two axes that were not directly tested in the main experiments: a different architecture family (Llama-3.1 vs.\ Qwen2.5) and a larger scale ($8$B vs.\ $1.5$B/$3$B). Together with the Qwen2.5 results in Table~\ref{tab:main_results}, this suggests the error-feedback mechanism is not backbone-specific and that its hyperparameters transfer across model families without retuning.

The scaling pattern is also consistent with the intuition that larger models are easier to optimize on a discrete lattice. With more redundant parameters and smoother loss landscapes~\cite{li2018visualizing}, the search for valid descent directions becomes more stable at the $8$B scale than at $1.5$B, where every grid step traverses a sharper region of the loss surface. This intuition matches the trend in Table~\ref{tab:main_results}, where QES gains grow with model size on GSM8K.

The memory implications are particularly relevant at this scale. Storing FP16 residuals for an $8$B model would require roughly $16$ GB of additional VRAM (Appendix~\ref{app:memory})---more than the quantized weights themselves and beyond the optimizer-state budget on most consumer GPUs. By contrast, QES holds the optimizer state to ${\sim}29.7$ KB regardless of model scale, making INT4 fine-tuning of an $8$B model feasible within the same memory envelope as inference.
\section{Detailed Ablation Results}
\label{app:ablations}

This appendix provides the full tables for the ablations summarized in Sections~\ref{subsec:ablation} and~\ref{subsec:fidelity}.
Table~\ref{tab:residual_ablation} compares Stateless Seed Replay against the Full Residual oracle on Countdown across all six (model, format) configurations.
Table~\ref{tab:window_decay} reports the effect of the replay window $K$ and decay factor $\gamma$ (top) and the update- and boundary-hit ratios used in the fidelity analysis (bottom).

\begin{table}[ht]
\centering
\caption{Countdown accuracy (\%): Stateless Seed Replay (QES) vs.\ Full Residual oracle. The seed-replay variant tracks the oracle within a few points across all configurations while reducing optimizer-state memory from gigabytes to kilobytes (Appendix~\ref{app:memory}).}
\label{tab:residual_ablation}
\vskip 0.05in
\begin{small}
\begin{sc}
\begin{tabular}{llcc}
\toprule
\textbf{Model} & \textbf{Format} & \textbf{QES} & \textbf{Full Residual} \\
\midrule
\multirow{3}{*}{Qwen2.5-1.5B}
 & INT4 & 16.00 & 18.05 \\
 & INT8 & 26.35 & 22.10 \\
 & W8A8 & 15.35 & 15.25 \\
\midrule
\multirow{3}{*}{Qwen2.5-3B}
 & INT4 & 31.85 & 33.50 \\
 & INT8 & 37.40 & 33.30 \\
 & W8A8 & 21.35 & 31.70 \\
\bottomrule
\end{tabular}
\end{sc}
\end{small}
\end{table}

\begin{table}[ht]
\centering
\caption{Seed-replay ablations on Qwen2.5-1.5B INT4 (Countdown). \textbf{Top:} Replay window $K$ and decay $\gamma$; \emph{Scaled} sets $\gamma^K\!\approx\!0$, \emph{Fixed} uses $\gamma{=}0.90$. \textbf{Bottom:} Update ratio and boundary-hit ratio $\rho$ in Stateless Seed Replay; reconstruction errors require both to coincide and are therefore vanishingly rare.}
\label{tab:window_decay}
\vskip 0.05in
\begin{small}
\begin{sc}
\begin{tabular}{ccc}
    \toprule
    \textbf{Window $K$} & \textbf{Decay $\gamma$} & \textbf{Accuracy (\%)} \\
    \midrule
    \multicolumn{3}{c}{\emph{Scaled decay}} \\
    50 & 0.90 & 16.00 \\
    40 & 0.87 & 14.00 \\
    30 & 0.83 & 14.15 \\
    20 & 0.78 & 10.80 \\
    10 & 0.58 & 4.55  \\
    \midrule
    \multicolumn{3}{c}{\emph{Fixed decay}} \\
    50 & 0.90 & 16.00 \\
    40 & 0.90 & 14.80 \\
    30 & 0.90 & 16.15 \\
    20 & 0.90 & 14.75 \\
    10 & 0.90 & 13.05 \\
    \bottomrule
\end{tabular}

\vskip 1em

\begin{tabular}{lcc}
\toprule
\textbf{Quantization} & \textbf{Update Ratio} & \textbf{Hit Ratio ($\rho$)} \\
\midrule
INT4 & $\approx 10^{-2}$ & $< 1 \times 10^{-5}$ \\
INT8 & $\approx 10^{-2}$ & $\approx 1.4 \times 10^{-4}$ \\
W8A8 & $\approx 10^{-2}$ & $\approx 6 \times 10^{-6}$ \\
\bottomrule
\end{tabular}
\end{sc}
\end{small}
\end{table}
\section{Memory and Compute Footprint}
\label{app:memory}

This appendix details the peak VRAM and wall-clock measurements summarized in Section~\ref{sec:experiments}.

\subsection{Memory}
\label{app:memory:vram}

Table~\ref{tab:memory} reports peak VRAM for the model weights and the optimizer state across methods. QES holds optimizer state to a constant ${\approx}29.7$~KB across all configurations---the size of the seed-and-reward history buffer---and therefore matches QuZO's total footprint, which is itself the inference-only baseline. Full Residual, by contrast, must materialize FP16 residuals for every parameter, adding $2.44$~GB for the $1.5$B model and $5.17$~GB for $3$B and effectively undoing the memory benefit of quantization. For reference, standard Quantization-Aware Training (QAT) on a $3$B model requires over $24$~GB of VRAM for full-precision weights, gradients, activations, and optimizer states---roughly $13\times$ the QES footprint at INT4.

\begin{table}[!t]
\centering
\caption{Peak VRAM on Qwen2.5. QES matches QuZO's inference-only footprint (optimizer state ${\approx}29.7$~KB across all settings); Full Residual adds gigabytes of FP16 buffers.}
\label{tab:memory}
\vskip 0.05in
\begin{small}
\begin{sc}
\begin{tabular}{llcccc}
\toprule
& & \textbf{Model} & \multicolumn{3}{c}{\textbf{Total VRAM (GB)}} \\
\cmidrule(lr){4-6}
\textbf{Model} & \textbf{Fmt.} & \textbf{Wts.\ (GB)} & QuZO & Full Res. & QES \\
\midrule
\multirow{3}{*}{Qwen2.5-1.5B}
 & INT4 & 1.071 & 1.071 & 3.511 & \textbf{1.071} \\
 & INT8 & 1.686 & 1.686 & 4.126 & \textbf{1.686} \\
 & W8A8 & 2.091 & 2.091 & 4.532 & \textbf{2.091} \\
\midrule
\multirow{3}{*}{Qwen2.5-3B}
 & INT4 & 1.926 & 1.926 & 7.094 & \textbf{1.926} \\
 & INT8 & 3.228 & 3.228 & 8.396 & \textbf{3.228} \\
 & W8A8 & 3.746 & 3.746 & 8.914 & \textbf{3.746} \\
\bottomrule
\end{tabular}
\end{sc}
\end{small}
\end{table}

\subsection{Wall-clock Time}
\label{app:memory:time}

Stateless Seed Replay introduces a reconstruction step whose cost scales linearly with the window $K$. Table~\ref{tab:walltime} reports per-iteration measurements on A100 GPUs at $K{=}50$.

\begin{table}[!t]
\centering
\caption{Per-iteration wall-clock time on A100 GPUs (Qwen2.5, INT4, $K{=}50$). Rollouts are run on 4 GPUs; replay on 1 GPU.}
\label{tab:walltime}
\vskip 0.05in
\begin{small}
\begin{sc}
\begin{tabular}{lcc}
\toprule
\textbf{Model} & \textbf{Rollout (s)} & \textbf{Replay (s)} \\
\midrule
Qwen2.5-1.5B + INT4 & 419  & 280 \\
Qwen2.5-3B + INT4   & 1017 & 522 \\
\bottomrule
\end{tabular}
\end{sc}
\end{small}
\end{table}

Under equal hardware budgets---i.e., rollouts and replay using the same number of GPUs (as an estimation, 419s and 1017s using 4 GPUs are equivalent to 1676s and 4068s using 1 GPU for rollouts)---replay adds ${\sim}16.7\%$ (1.5B) and ${\sim}12.5\%$ (3B) overhead over the no-replay oracle. This is the worst case for a naive sequential implementation and can be reduced through two practical levers:

\begin{itemize}
    \item \textbf{Tunable window size.} As shown in Section~\ref{subsec:ablation}, $K{=}20$ retains most of the accuracy of $K{=}50$ at $40\%$ of the reconstruction cost; performance degrades gracefully as $K$ decreases when the decay $\gamma$ is held constant.
    \item \textbf{Parallel reconstruction.} Because no inference traffic is served during the update phase, the model's pre-allocated KV-cache memory and other idle resources are available. Reconstruction across layers can be parallelized using this otherwise-unused memory, removing the strict sequential dependency that drives the worst-case figure above.
\end{itemize}

The actual penalty is therefore highly dependent on the inference engine and target hardware rather than being a strict linear multiplier on training time.
\section{Continuous Optimization in the Discrete Grid}
\label{app:continuous}
\begin{figure}[ht]
    \centering
    \includegraphics[width=0.4\linewidth]{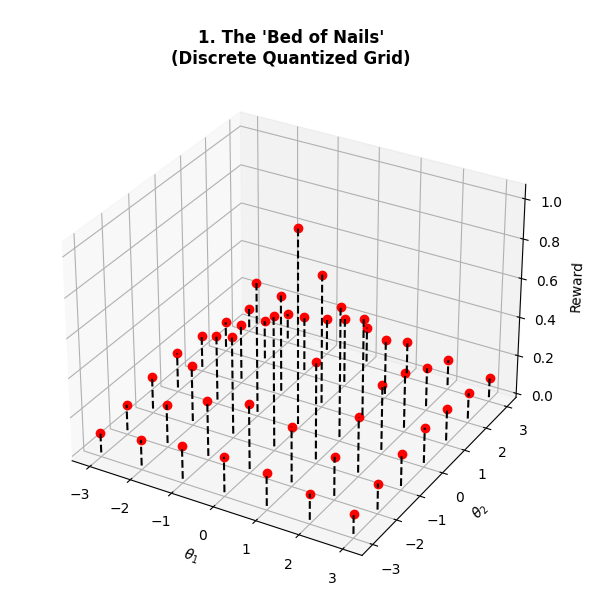}
    \hfill
    \includegraphics[width=0.4\linewidth]{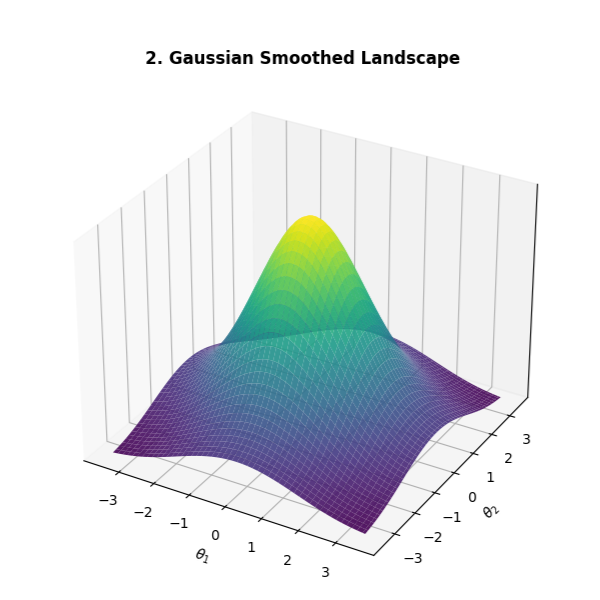}
     \caption{A continuous reward function and its instantiation on a discrete grid. While the reward function can be optimized with a high-precision gradient ascent, that method cannot be applied to the discrete case. Instead, the gradient needs to be estimated without vanishing signals or inaccurate moves between grid points. That is the challenge that QES is designed to solve.}
    \label{fig:illustrate_discrete}
\end{figure}

\section{Broader Impacts} \label{app:impact}

This paper presents Quantized Evolution Strategies (QES), an optimization framework designed to democratize access to Large Language Model (LLM) fine-tuning. By enabling high-precision learning directly within a quantized parameter space, QES allows full-parameter adaptation on consumer-grade hardware that was previously restricted to static inference. Furthermore, the significant reduction in memory overhead contributes to more resource-efficient AI development, potentially lowering the energy footprint and environmental impact of scaling large models.



\end{document}